# SPEECH CONTROLLED QUADRUPED

Devashish Kulkarni

*Department of electronics and telecom*

*Xavier Institute of Engineering*

*Mahim, Mumbai 400037, India*

devashishkulkarni1994@gmail.com

Sagar Paldhe

*Department of electronics and telecom*

*Xavier Institute of Engineering*

*Mahim, Mumbai 400037, India*

Sagar.paldhe@gmail.com

***Abstract-*** The project which we have performed is based on voice recognition and we desire to create a four legged robot that can acknowledge the given instructions which are given through vocal commands and perform the tasks. The main processing unit of the robot will be Arduino Uno. We are using 8 servos for the movement of its legs while two servos will be required for each leg. The interface between a human and the robot is generated through Python programming and Eclipse software and it is implemented by using Bluetooth module HC 06.

## 1. INTRODUCTION

The concept of creating machines that can operate autonomously dates back to classical times, but research into the functionality and potential uses of robots did not grow substantially until the 20th century. Today, robotics is a rapidly growing field, as technological advances continue; researching, designing, and building new robots serve various practical purposes, whether domestically, commercially, or militarily. Many robots do jobs that are hazardous to people such as defusing bombs, mines and exploring shipwrecks. Walking is a difficult and dynamic problem to solve; we have overcome this problem by using a pair of servos for each leg of this quadruped. The dimensions of all the four legs as well as the abdominal region of the robot were not available to us. We studied the servos that we had bought beforehand to make the body parts using acrylic sheet and laser cutting.    The size of the servos helped us a lot to constitute the different parts. After designing the legs and the abdomen using Coral Draw we made the parts using acrylic sheet. We also made a copy of the legs prior to this using normal cardboard so as to make sure that the dimensions were right and that everything fitted properly. When the parts were cut we used a standard adhesive to join both the sections of the legs together. We then studied Arduino Programming and developed the program for the quadruped using the Integrated Development Environment (IDE) and burned it into the Arduino. Male to Female wires were used to make the connections between the servos and the pins of the Arduino. After countless experiments with the program we were able to implement the final product. The interfacing is provided by using Python programming and Eclipse. The interface between the robot and the mobile is provided by the Bluetooth application in the PC and the Bluetooth Module HC 06, hence both of them are connected wirelessly.

## 2. COMPONENTS

### 2.1 HARDWARE REQUIRED

1. ARDUINO UNO

 Arduino is an open-source physical computing platform which has a simple I/O board and the programs are implemented by using an Integrated Development Environment (IDE) for Arduino. The Arduino Uno has a microcontroller named ATmega328 (datasheet). It has 14 digital input/output pins (of which 6 can be used as PWM outputs), 6 analog input pins, a 16 MHz ceramic resonator, a USB port, a power jack, an ICSP header, and a reset button. It contains everything needed to support the microcontroller ATmega328 which is the core of the Arduino; we can connect it to a computer with a USB cable or power it with an AC-to-DC adapter or battery for it to start.

 **Summary**

Microcontroller        ATmega328

| | |
|---|---|
| Operating Voltage | 5V |
| Input Voltage (recommended) | 7-12V |
| Input Voltage (limits) | 6-20V |
| Digital I/O Pins | 14 ( 6 provide PWM output) |
| Analog Input Pins | 6 |
| DC Current per I/O Pin | 40 mA |
| DC Current for 3.3V Pin | 50 mA |
| Flash Memory | 32 KB (ATmega328) of which 0.5 KB used by bootloader |
| SRAM | 2 KB (ATmega328) |
| EEPROM | 1 KB (ATmega328) |
| Clock Speed | 16 MHz |
| Length | 68.6 mm |
| Width | 53.4 mm |
| Weight | 25 g |

**Power**

The Arduino Uno can be powered by using the USB port which is connected to the computer or with an external power supply. The external power supply can be made by using a pair of pencil cells or by using a Lithium battery.

External (non-USB) power can also come from an AC-to-DC adapter. The adapter can be connected by plugging a 2.1mm centre-positive plug into the board's power jack. Leads from the battery can be connected to the VIN and GND pins of the Arduino.

The board can operate on an external supply voltage of about 6 to 20 volts. If the supplied voltage is less than 7V, however, the 5V pin may supply less than five volts and the board can become unstable. If more than 12V is applied, the voltage regulator may overheat and the Arduino board can get damaged. The recommended range is between 7V and 11V.

The power pins are as follows:

**GND.** Ground pins.

**VIN.** It is sometimes labelled as 9V. This is the input voltage to an Arduino board when we are using an external power supply as opposed to the 5V which we get from the USB connection.

**5V.** This pin gives an output of 5V which is regulated from the regulator on the board. If voltage is supplied through the 5V or 3.3V supply pins, bypassing the regulator, the board can get damaged so it is not advisable.

**3V3.** A 3.3 volt supply generated is by the regulator present on the board. The maximum current draw is close to 50 mA.

2. BLUETOOTH MODULE HC-06

HC-06 module is a Bluetooth SPP (Serial Port Protocol) module which is designed to established connection wirelessly.

The range of Bluetooth module HC 06 is about 5 meters and it can be easily connected to other Bluetooth modules like the ones used PCs and mobile phones within that radius.

**Hardware Features**

- Typical -80dBm sensitivity
- Up to +4dBm RF transmit power
- Low Power 1.8V Operation ,1.8 to 3.6V needed for I/O operations
- PIO control
- UART interface with programmable baud rate
- Integrated antenna is available
- Edge connector is available

**Software Features**

- Default Baud rate: 38400, Data bits: 8, Stop bit: 1, Parity: No parity
- Baud rate:9600,19200,38400,57600

- Given a rising pulse in PIO0, device will be disconnected.
- Status instruction port PIO1: low-disconnected, high-connected

- Auto-pairing PINCODE:"0000" as default

3. SERVOS

Servo or RC **Servo Motors** are DC motors which are used for precise control of angular position. The RC servo motors usually have a rotation limit from 90° to 180°. Some **servos** even have a rotation limit of 360°.

A **servomotor** is a rotary actuator which allows precise control over angular position, velocity and acceleration. Servos are controlled by sending an electrical pulse of variable width, or which is otherwise known as pulse width modulation (PWM), through the controlling wire. There is a minimum pulse, a maximum pulse, and a repetition rate. A servo motor can turn for about 90 degrees in either direction which accounts for a total of 180 degree movement. The PWM signal which is sent to the motor controls position of the shaft, and according to the duration of the pulse sent via the control wire; the rotor will turn to the desired position. The servo motor receives a pulse every 20 milliseconds (ms) and the length of the pulse determines the rotary action of the motor. The servo motors are used for the hip and knee movements of the robot and thus it helps it in performing different actions.

DATA SHEET:

Operating Voltage : 4.8-6.0V

PWM Input Range : Pulse Cycle 20±2ms, Positive Pulse 1~2ms

Stall Torque : 0.8 kg at 4.8V

Weight : 9 g

Size : 22*12.5*20*26.6

4. ACRYLIC SHEET

The Acrylic sheet is made of Poly (methyl methacrylate) (PMMA) which is a transparent thermoplastic often used as a lightweight and a shatter-proof alternative to soda-lime glass. PMMA is generally produced by emulsion polymerization, solution polymerization, and bulk polymerization. To produce 1 kg (2.2 lb) of PMMA, about 2 kg (4.4 lb) of petroleum is needed. PMMA produced by radical polymerization is completely amorphous. We have used the acrylic sheets for making the skeleton of the robot to make its legs and abdomen.

5. JUMPER WIRES

We used male to female jumper wires for this project. They were used to connect the servos to the pins of the Arduino Uno. The male part of the wire was inserted in the servo while the female pare was connected to the pins of the Arduino Uno. They are used to transfer electrical signals from anywhere on the breadboard to the input/output pins of a microcontroller. We required 40 of these male to female wires to make all the connections.

6. USB CABLE

Universal Serial Bus (USB) is an industry standard cable developed in the mid-1990s. It is only used to connect the board of the robot to the computer for burning the software in the Arduino. It can be also used to power the Arduino by connecting it to the computer at one end and to the Arduino at the other end.

7. PENCIL CELLS

Primary zinc-carbon batteries are used as an external power supply. Each zinc-carbon battery or otherwise known as pencil cell produces a maximum voltage of 1.5V so we have used a pack of 4 pencil cells which is attached to the lower side of the abdominal region of the robot. The positive terminal of the pack is connected to VIN and the negative terminal is connected to GND pin of the board.

8. BREADBOARD

The breadboard is used to provide the 5V and GND to the servo motors. The 5V and GND supply pin of the Arduino is connected to a point on the breadboard and it is then provided to the servo motors.

2.2: SOFTWARE REQUIRED

1. ARDUINO UNO PROGRAMME

For programming the microcontrollers, the Arduino platform provides an integrated development environment (IDE) which includes support for C and C++ programming languages. The Arduino integrated development environment (IDE) is a cross-platform application written in Java. It is basically used to introduce programming to artists and other newcomers unfamiliar with software development. It is also capable of compiling and uploading programs to the board with a single click. A program or code written for Arduino is known as a sketch.

2. PYTHON 3.4

Python is a widely used general-purpose, high-level programming language which can be used to implement a large number of applications. Its syntax allows programmers to express concepts in fewer lines of code than through languages such as C++ or Java.

We are using Python version 3.4 for implementing the voice control mechanism. Bitvoicer is another program that can be used instead of Python. Python is itself implemented by using a programme known as Eclipse or Geany. It will be used to recognise the vocal commands given by the user interpret it and then pass on the information to the quadruped.

Note that we are using version 3.4 of Python since the lower versions do not contain the speech recognition format.

### 3. ECLIPSE

In computer programming, Eclipse is an integrated development environment (IDE). It is written mostly in Java and is used to develop different applications.

We are going to use Eclipse as an interface to run the Python programme. The other alternative for Eclipse is Geany which is another type of IDE; but as Eclipse is more robust we decided to go with Eclipse.

### 4. COREL DRAW

We have used Corel Draw to prepare the dimensions for the quadruped. Programs like AutoCAD can also be used in place of it.

## 4. WORKING

### 3.1 HARDWARE ASSEMBLY

The hardware components which are present in this quadruped consist of the Arduino Uno, the breadboard, the 8 servos, acrylic parts, connecting wires and the power supply.

### 1. PARTS MADE FROM ACRYLIC SHEET

The four legs and the abdomen of the quadruped are made up of acrylic sheet. The dimensions were obtained by using the servos as a reference. Coral draw X was used to design both the parts of the legs as well as the abdomen of the quadruped. The parts were cut-out by using a technique known as 'Laser Cutting'. Both the separate parts of the legs were joined together by using an Adhesive. Cavities and holes of different shapes and dimensions are inscribed in the parts for the fitting of the Arduino and the servos.

### 2. SERVO ORIENTATIONS

Step 1: Connect a servo to the Arduino Uno by connecting the Red wire of the servo to the +5V pin of the Arduino, the Brown wire to the ground and the Orange wire to pin number 9 for giving the digital input.

Step 2: Connect the Arduino to the computer and go to the IDE. Then go to examples and run the sweep program after making proper changes in it as shown below so that the servo motor is fixed at 90 degree angle orientation. The only change that is needed is to make the angle equal to 90 Degrees from 180 degrees. The example is given below.

```
void setup()
{
  myservo.attach(9);  // attaches the servo on pin 9 to the servo object
}

void loop()
{
  for(pos = 0; pos < 90; pos += 1)  // goes from 0 degrees to 180 degrees
  {                      // in steps of 1 degree
    myservo.write(pos);              // tell servo to go to position in variable 'pos'
    delay(15);                       // waits 15ms for the servo to reach the position
  }
}
```

Step 3: Perform the above given procedure for all the 8 servos which are going to be used. Now the next step is to attach the servos to the legs of the quadruped.

### 3. SERVO ASSEMBLY

Step 1: Attach the foam rubber feet to the legs as shown using 3mm x 12mm pan head screws.

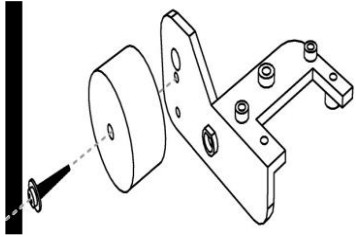

Step 2: Attach a servo on each of the leg segments as shown using 2.3mm x 8mm self-tapping screws. We have also used a general adhesive to attach the servos to the acrylic sheet so that they remain stable.

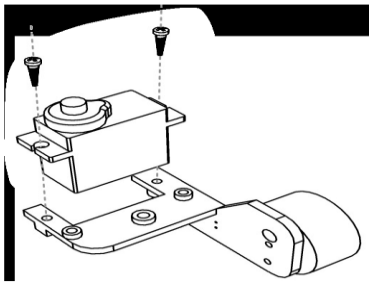

Step 3: Mount a servo on to an unused leg segment as shown by using 2mm x 6mm self-tapping screws. Orient the servo to the neutral position and then fit the leg segment to the servo using a 2mm x 8mmpan head screw as shown. Gently turn the servo by hand to check the range of movement and adjust if necessary.

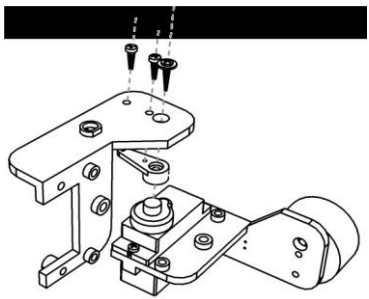

Step 4: Now we should make two legs for the right side of the chassis. Mount two servos as shown in the diagram with 2.3mm x 8mm self-tapping screws. Note the orientation of the servoas shown in the figure.

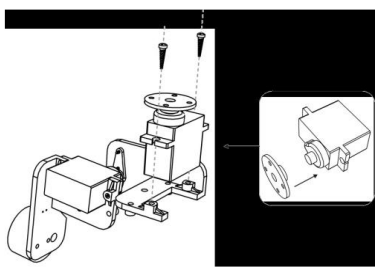

Step 5: We make two legs for the left side of the robot in a similar manner as step 4 except that the servo is mounted the opposite way.

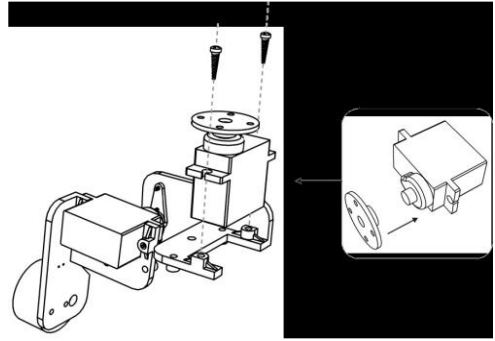

Step 6: The orientation of the servo should be in the centre and mount the legs onto the mounting plate using 2mm x 6mm pan head screws. Use a 2mm x 8mm pan head screw in the centre of the servo horn to secure the servo to the mounting plate. When mounting the servo horn to the mounting plate the holes marked in red are recommended. Gently move the legs by hand to check their range of movement and adjust the servo horns as required. Your chassis is now ready to use.

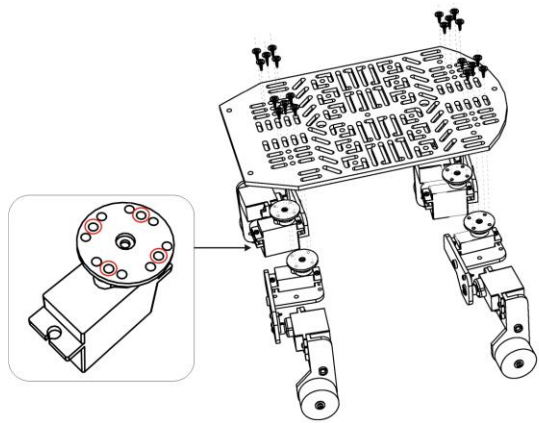

4. SERVO CONNECTIONS

To provide the digital input to the servos connect the Orange wires to the digital output pins of the Arduino in the following manner. We can use male to female jumper cables to make the connection.

| SERVOS | ARDUINO PINS |
|---|---|
| Front left hip | 11 |
| Front right hip | 10 |
| Rear left hip | 9 |
| Rear right hip | 8 |
| Front left knee | 6 |
| Front right knee | 5 |
| Rear left knee | 3 |
| Rear right knee | 4 |

Connect all the Red wires of the servos to the +5V VCC pin of the Arduino and the Brown wires to the GND pin using a bread board or a PCB board.

5. POWER SUPPLY

The board can operate on an external supply of 6 to 20 volts. If supplied with less than 7V, however, the 5V pin may supply less than five volts and the board may be unstable. If using more than 12V, the voltage regulator may overheat and damage the board. The recommended range is 7 to 12 volts.

6. BLUETOOTH MODULE CONNECTIONS

To connect the Bluetooth module HC 06 to the Arduino Uno make the following connections.

 GND - GND

 VCC - +5V

 TXD - RX0 / D0 (Signal)

 RXD - TX0 / D1 (Signal)

To connect the Bluetooth module of your PC to the one in the quadruped perform the following steps.

1. Right Click on Bluetooth Devices icon in the tray bar and click add new device.

2. Let it search until HC - 06 pops up.

3. Double click it.
In the next dialogue box select enter the device's pairing code and type in 1234. It should then install the HC 06 to your computer.

4. Now this part is important open up device manager (Start >> Device Manager)

Go to Ports (COM & LPT) and note down the C

3.2 SOFTWARE ASSEMBLY

The software which is required for this project is the Arduino Uno program, IDE for burning the Arduino program, Python programming for speech controlling and Eclipse for burning the python program.

1. BURNING THE PROGRAM INTO ARDUINO UNO

Connect the Arduino Uno to the computer using a USB cable, open IDE and perform the following steps.

Step 1: Open device manager and look up Ports (COM & LPT).

Step 2: Tools >> Board >> Arduino Uno.

Step 3: Tools >> Processor >> Arduino Uno

Step 4: Tools >> Port >> COM3

Step 5: Now hit the upload button.

2. SPEECH RECOGNITION USING PYTHON AND ECLIPSE

We are using Python programming for speech recognition. Perform the following steps to burn the program given in chapter 2.

Step 1: Download Python 3.4 and save it as C:\Python34.

Step 2: Download and install eclipse which is an interface for writing your code via Python.

Step 3: Install the libraries which we need for speech recognition.

Libraries Required:

PyBluez 0.20 - Bluetooth (setup)

PyAudio 0.2.8 - provides bindings for speech recognition (setup)

SpeechRecognition 1.1.0 - speech recognition (download)

For installing them open and download the libraries. Download "SpeechRecognition-1.1.0.tar.gz (md5)". Extract all the contents of the downloaded folder via winRAR to the python 3.4 directories (C:\Python34). If a dialogue box appears asking whether to replace existing files, click yes to all. Now navigate to C: drive and while holding Shift, right click on the folder that says Python34 and click "Open command window here". Type in "python setup.py install" without quotes and hit Enter. Now you should have all the required libraries installed.

Step 4: Open eclipse.

Step 5: Copy the saved Python program and paste it in eclipse.

Step 6: Right click on the Bluetooth icon on the tray and click on "Show Bluetooth Devices". Right click on HC-06 and click on properties. Go to the Bluetooth tab and copy the Unique identifier number and paste that in the place of xx: xx: xx: xx: xx: xx.

## 5. CONCLUSION

We have successfully demonstrated the concept of voice controlled robots through this experiment as well as we have done it in an affordable manner, the cost of our entire project is Rs 3500. This is merely a demonstration of technology that we can control the robotic movements through vocal commands; more advancement can be done in this by using more number of servos and a more powerful processor. The speech controlled quadruped has been made by using software like Arduino programming, Python and Eclipse which are open source due to which we were easily able to make the necessary programs and no money was wasted to acquire the software. The Arduino is also non-propriety in nature so it is available for an affordable cost of Rs 400 which, so we have also demonstrated that we can make such complex robots by using components which are freely available. The decision to perform this project was mainly because it is a good example of an embedded system. This robot can be made by using different designs according to the availability of other components and resources. This design was envisioned by us and it is not a compulsion that the design has to be in this manner hence different versions of this robot can be made easily according to its applications.

## 5. FUTURE SCOPE

There are a lot of variations that can be done on the speech controlled quadruped. It can be upgraded easily by using powerful processors and servo motors. The design of this robot can be varied according to its applications and there are a lot of applications in which it can be used. An Obstacle sensor can be integrated in this design by using ultrasonic or infrared sensors in the front or back region. Suction cups can also be connected to its legs using required mechanism so that it can have a better grip on the surface that it is walking on so that it can easily climb inclined surfaces and maybe even on walls. The speech controlled quadruped has limitless potential and can be used for varied applications for industrial to commercial and situations by altering certain characteristics and mechanisms. By upgrading it in a proper manner we can customise the speech controlled quadruped to go to certain places which are hazardous in nature like volcanoes, caves and for deep ocean research. It can also be controlled easily through vocal commands and told to do necessary actions. Various types of sensors can be mounted on it can be used to gather data in a variety of ecosystems and even in deep space; hence it can be used for research operations. Many other instruments and extensions can be attached to it like a robotic arm so that it can be used for military operations like in diffusion of a bomb or in accidents where it is difficult for the personnel to reach the victims. It can also be used for household applications like cleaning etc. This is very important because anyone can make it using their own designs and customise it for their different applications. The possibilities for it are endless and it can be implemented in various situations for the betterment of mankind.

## 6. REFERENCES

[www.wikipedia.com](www.wikipedia.com)

[www.eclipse.en](www.eclipse.en)

[www.pythonhelpdesk.en](www.pythonhelpdesk.en)

[www.bluetooth.us](www.bluetooth.us)

[www.arduino.com](www.arduino.com)

[www.robotgear.com](www.robotgear.com)